\documentclass{bmcart}
\usepackage{graphicx}
\usepackage{multirow}
\usepackage[utf8]{inputenc} 

\startlocaldefs
\endlocaldefs

\begin{document}

\begin{frontmatter}

\begin{fmbox}
\dochead{Research}


\title{Bi-Encoders based Species Normalization - Pairwise Sentence Learning to Rank}

\author[
addressref={aff1},                   
corref={aff1},                       
email={zainabkhalid.awan@alumni.uts.edu.au}   
]{\inits{Z}\fnm{Zainab} \snm{Awan}}
\author[
addressref={aff2},
email={tim.kahlke@uts.edu.au}
]{\inits{TK}\fnm{Tim} \snm{Kahlke}}
\author[
addressref={aff2},
email={peter.ralph@uts.edu.au}
]{\inits{PJ R}\fnm{Peter J.} \snm{Ralph}}
\author[
addressref={aff1},
email={paul.kennedy@uts.edu.au}
]{\inits{PJ K}\fnm{Paul J.} \snm{Kennedy}}

\address[id=aff1]{
	\orgname{Australian Artificial Intelligence Institute, UTS}, 
	\street{NSW},                     %
	\city{Sydney},                              
	\cny{Australia}                                    
}
\address[id=aff2]{%
	\orgname{Climate Change Cluster, UTS},
	\street{NSW},
	\city{Sydney},
	\cny{Australia}
}

%


\begin{artnotes}
\end{artnotes}

\end{fmbox}


\begin{abstractbox}

\begin{abstract} 
\parttitle{Motivation} 
Biomedical named-entity normalization involves connecting biomedical entities with distinct database identifiers in order to facilitate data integration across various fields of biology. Existing systems for biomedical named entity normalization heavily rely on dictionaries, manually created rules, and high-quality representative features such as lexical or morphological characteristics. However, recent research has investigated the use of neural network-based models to reduce dependence on dictionaries, manually crafted rules, and features. Despite these advancements, the performance of these models is still limited due to the lack of sufficiently large training datasets. These models have a tendency to overfit small training corpora and exhibit poor generalization when faced with previously unseen entities, necessitating the redesign of rules and features.

\parttitle{Contribution} 
We present a novel deep learning approach for named entity normalization, treating it as a pair-wise learning to rank problem. Our method utilizes the widely-used information retrieval algorithm Best Matching 25 to generate candidate concepts, followed by the application of bi-directional encoder representation from the encoder (BERT) to re-rank the candidate list. Notably, our approach eliminates the need for feature-engineering or rule creation. We conduct experiments on species entity types and evaluate our method against state-of-the-art techniques using LINNAEUS and S800 biomedical corpora. Our proposed approach surpasses existing methods in linking entities to the NCBI taxonomy. To the best of our knowledge, there is no existing neural network-based approach for species normalization in the literature."
\end{abstract}




\end{abstractbox}
%

\end{frontmatter}



\section*{Introduction}
Biomedical named-entity normalization (NEN) is the process of assigning a unique identifier to a biomedical entity. For example, \textit{homo sapien} or a human is a species that is assigned the identifier 9606 in the NCBI taxonomy\footnote{https://www.ncbi.nlm.nih.gov/taxonomy/?term=human}.
Biomedical named-entity recognition identifies the type of a biomedical entity, such as gene, protein, cell-line, disease, chemical, drug or species. Once identified, an entity has to be linked with a standard knowledge base such as the NCBI taxonomy \cite{schoch2020ncbi} for species, PubChem for chemicals, Gene Ontology \cite{carbon2021gene} for genes, UniProt \cite{apweiler2004uniprot} for proteins and the CTD (comparative toxicogenomics database) \cite{davis2021comparative} for diseases. In the literature, NEN is also referred to as entity linking or entity disambiguation. Entity normalization is a primary task in any biomedical information extraction pipeline after named entity recognition (NER) which is defined as linking named entities to a standard database identifier and before relation extraction \cite{yadav2018survey} and knowledge base construction \cite{szklarczyk2016string}. Entity linking is essential to such a pipeline as its applications include semantic web, information retrieval, knowledge base construction and recommender systems, to name a few.

Entity linking poses challenges including, ambiguity and term variation. Ambiguity is when an entity can be linked to more than one identifiers, for example, \texttt{perennis} can be linked to the taxa \texttt{Bellis perennis} or \texttt{Monopera perennis} with the taxonomy IDs $41492$ or $2584082$ respectively depending on the context. ``Term variation" is when multiple syntactically different terms could be mapped to the same identifier. For instance, two morphologically different terms \texttt{Drosophila melanogaster} and \texttt{fly} are mapped to the same unique identifier $7227$ in the NCBI taxonomy. Term variation is a major challenge in biomedical entity linking, whereas ambiguity is more prevalent in the general domain. 

In this paper, we investigate the issue of species named entity normalization, wherein a method is employed to assign a distinct identifier from the NCBI taxonomy to a given set of species. The recognition of species and their linkage to the NCBI taxonomy hold significant significance as they aid scientists in species identification, computational knowledge inference, and extraction.

Our proposed approach draws inspiration from a recently introduced model in the field of medical entity normalization, aiming to reduce the reliance on feature engineering and rule creation \cite{ji2020bert}.   
In the next section we discuss related work \ref{relatedwork}, then our proposed method\ref{proposedmethod}. Next we describe the experimental setup\ref{experimentalsetup}, results and discussions\ref{resultsanddiscussions} and finally we conclude \ref{conclusion1} our paper.

\subsection*{Contribution}
We apply BERT-based ranking to species normalization to the NCBI taxonomy and benchmark our approach on LINNAEUS and S800 corpus. We compare our approach with two competitive baselines and demonstrate that those baseline methods cannot capture species' semantics and hence underperform. 

\section*{Related Work}\label{relatedwork}
Biomedical named-entity normalization has been approached with dictionary-based as well as machine learning-based approaches. Also, most of the existing works for species focus on species recognition and normalization (linking to the NCBI) as a subsequent step. 

For chemical entity normalization tmChem \cite{leaman2015tmchem} utilize a dictionary-based approach after performing chemical entity recognition. The chemical entities were collected from MeSH \footnote{https://www.nlm.nih.gov/mesh/meshhome.html} and ChEBI\footnote{https://www.ebi.ac.uk/chebi/}. Sieve-based entity linking \cite{d2015sieve} is an approach that uses ten sieves or rules to assign an identifier to disorder mentions. Sieves are rules such as an exact match or partial match. If a disorder mention does not pass any sieve, it is assigned the identifier ``unlinkable". OrganismTagger \cite{naderi2011organismtagger} is a rule-based system for recognizing and normalizing organisms to the NCBI identifiers. Dictionary-based methods require the dictionaries to be comprehensive and re-processed every time a new organism is added to the database. Hence, speed is a concern for dictionary-based approaches.

Kate \cite{kate2016normalizing} proposed an edit distance-based method for disease and disorder mention normalization, capable of automatically learning term variations/rules. The method, however, does not understand the semantics of mentions. 

Li and colleagues \cite{li2017cnn} apply a convolutional neural network (CNN) based network to disease normalization which uses a sieve-based approach \cite{d2015sieve} to generate candidate concepts for each disease mention and then maximize the similarity between mention and candidate concept pair. Cho et al. \cite{cho2017method} use word2vec embeddings to map the plant entity mentions and concepts form knowledge bases on vector space and maximize the cosine similarity between the pair for diseases and plant names. Kaewphan et al. \cite{kaewphan2018wide} propose a normalization method for multiple biomedical entity types based on fuzzy string matching. Entity mention and concept are mapped to vectors using character n-gram frequencies and cosine similarity is used as a scoring function to maximize the similarity between entity mentions and concept terms. DNorm \cite{leaman2013dnorm} applies pair-wise learning to rank approach to the disease normalization task. The entities and concepts from knowledge base are represented as term frequency-inverse document frequency. A scoring function is learnt from training data that maximizes the similarity between entity mention and concepts.

Furrer et al. address normalization of the biomedical entities \cite{furrer2019uzh} as a sequence labelling problem using a bi-directional long short-term memory network (Bi-LSTM). However, in a sequence labelling problem, a model can normalize only those entities seen in the training set, which is a major limitation of such systems. 
Zhou et al. propose knowledge enhanced normalization for genes and proteins \cite{zhou2020knowledge}. Their method employs embedded language models (ELMo) and structural knowledge from NCBI and UniProt ontologies for protein and gene normalization with the performance of $44.5\%$ F1-measure.

NormCo \cite{wright2019normco} is a deep coherence model for disease normalization which combines and entity phrase model to capture a semantic model and a topical coherence model to learn about other disease mentions in a document. The final model is composed of combining these two models and are trained jointly. 
TaggerOne \cite{leaman2016taggerone} jointly trains for NER and NEN using semi Markov models. It uses a rich feature set for NER and tf-idf weights for NEN.
Deng and colleagues \cite{deng2019ensemble} propose a two-step convolutional neural network ensemble to normalize microbiology related entities.
Ferré et al. \cite{ferre2020c} propose a hybrid of word embeddings, ontological information and weak supervision method for normalizing Habitat and Phenotype entities to OntoBiotope ontology \cite{bossy2016ontobiotope}.

Recently, bidirectional encoder representation from Transformers (BERT) \cite{devlin2019bert} has been applied to biomedical/clinical entity normalization. BERT has been applied to clinical NEN in two different ways. Firstly, Li and colleagues \cite{li2019fine} apply BERT to electronic health records to normalize disorder mentions to knowledge bases by treating normalization as a token level task, and a number of disorder mentions in the knowledge base are the number of classes. Secondly, Ji and colleagues \cite{ji2020bert} apply BERT to disease NEN by approaching it as a ranking problem which is a sequence level task. They use the BM25 algorithm to generate candidate concepts and rerank the candidates by learning semantic similarity between entity and concept pairs.

\subsection*{Research Gaps}
Species normalization has mostly been addressed as a dictionary matching problem-as a subsequent step of named entity recognition. The issue with these approaches is that they do not capture the context and semantics of species names. The addition of a new species may require additional rules or features, and species may be linked to concepts that are lexically similar but semantically different. Here we apply `pair-wise learning to rank' using pre-trained BERT \cite{devlin2019bert} based language models for species normalization, which do not require any rules or features. The corpus can be easily constructed from the NCBI taxonomy with a simple script, and new entities can be added without changing rules or features. Although similar works have been done for the drugs and disease normalization, no work has been reported on linking taxonomic units to the NCBI identifiers. State-of-the-art NLP pipelines employ deep learning models for genes, disease, chemicals, drugs and mutations normalization and still rely on dictionary lookup approaches for species \cite{xu2020building}. There are no existing neural models for species normalization.

\section*{Baselines}\label{baselines}
\begin{center}
	\begin{figure}[h!] 
		\centering
		\includegraphics[scale=10]{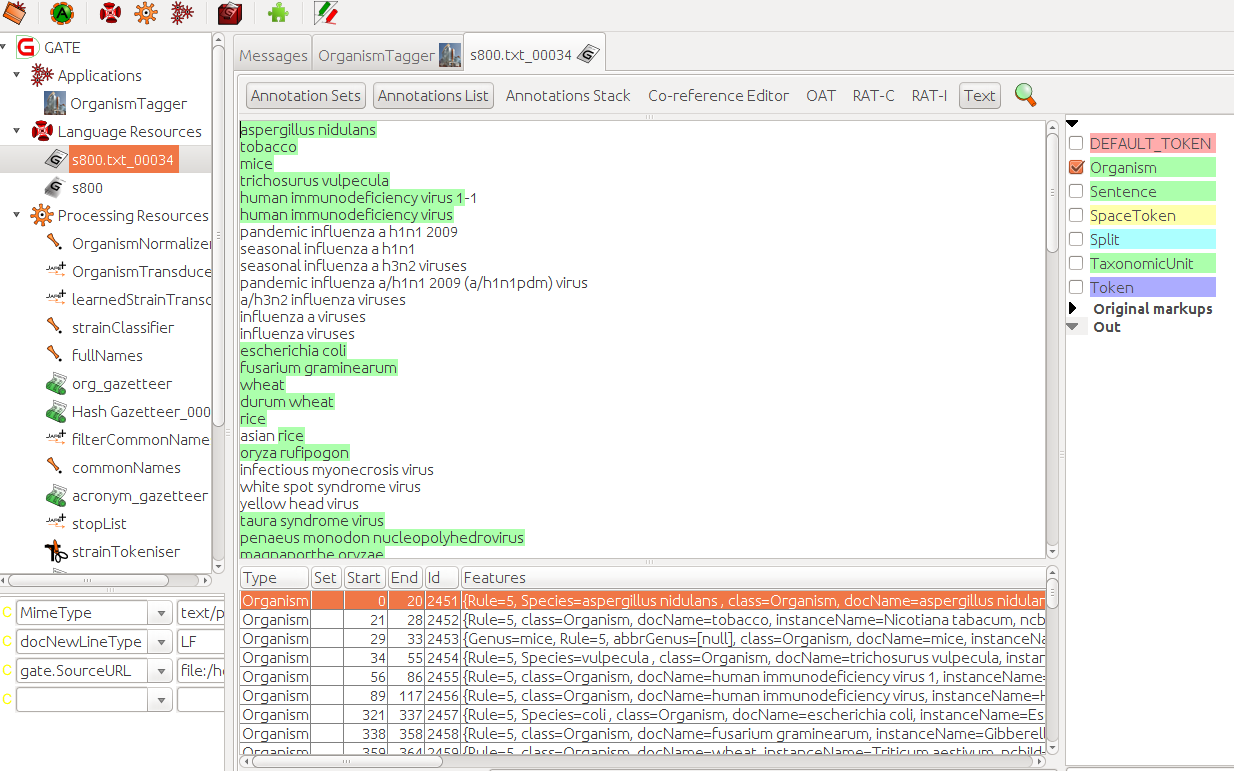}
		\caption{OrganismTagger - GATE based framework for organisms NER and NEN}
		\label{OT}
	\end{figure}
\end{center}

\begin{center}
	\begin{figure}[h!] 
		\centering
		\includegraphics[width=12cm, height=6cm]{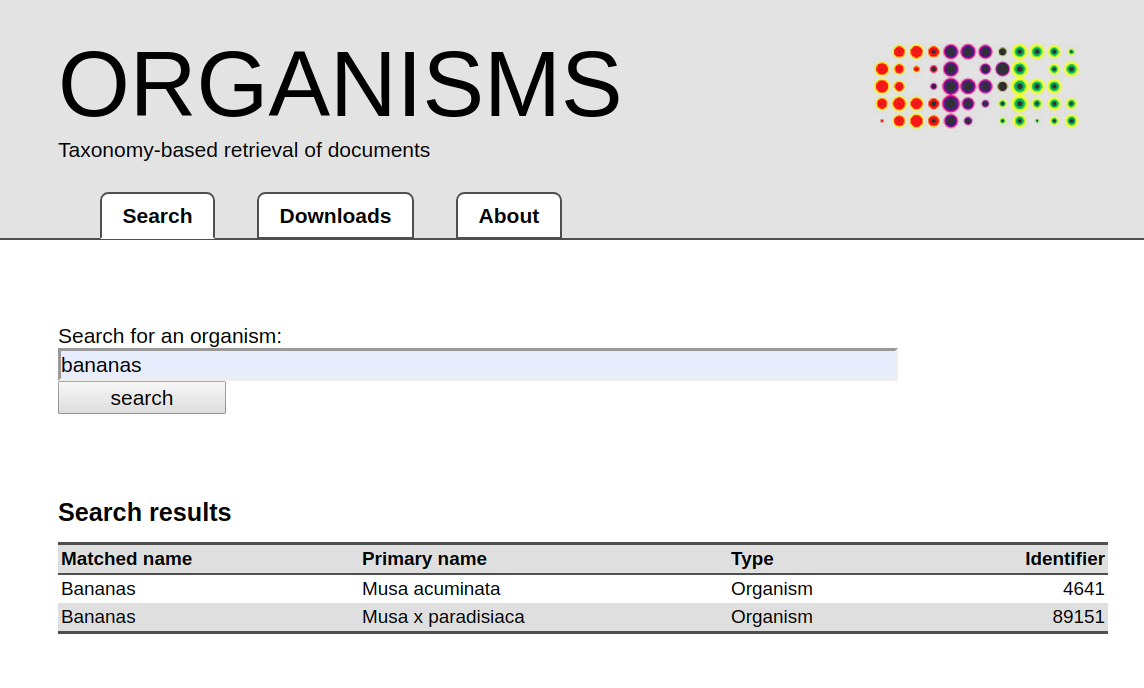}
		\caption{ORGANISMS - web-based resource for taxonomic names identification}
		\label{ORGANISMS}
	\end{figure}
\end{center}

We used OrganismTagger \cite{naderi2011organismtagger} and ORGANISMS web resource \cite{pafilis2013species} as baselines. OrganismTagger \cite{naderi2011organismtagger} is a hybrid of machine learning and rule-based techniques. It uses pre-processing modules from the General Architecture for Text Engineering (GATE) framework to pre-process text documents and the NCBI taxonomy database for assigning a unique identifier to identified mentions in documents. We used GATE developer\footnote{https://gate.ac.uk/family/developer.html} for tagging species with their unique identifier. In Figure \ref{OT},
we show the GUI of OrganismTagger, where species highlighted in green colour have been recognized and assigned a unique identifier. The species that are not highlighted have not been recognized and hence not linked with any identifier. 
ORGANISMS \cite{pafilis2013species} is a dictionary-based web resource for taxonomy-based retrieval of documents. It uses a dictionary that has taxonomic units from the NCBI taxonomy. For a given input mention, it retrieves a matched name, primary name, type and identifier. For instance, for the given query ``bananas", it retrieves two matches, out of which the first match is considered only as in Figure \ref{ORGANISMS}.

\section*{Proposed Method}\label{proposedmethod}
This section presents the problem definition and our proposed method to perform named entity normalization of species. 
\subsection*{Problem Definition}
We approach species normalization as a linking problem, where a given named entity $NE$ should be linked to a candidate concept (CC) from a knowledge base (KB), $\hbox{NE}  \rightarrow  \hbox{CC}$. The problem here is addressed as pair-wise learning to the rank problem (pLTR)\cite{liu2011learning}. In pLTR, named entities are considered queries and candidate concepts from the knowledge base form pairs with queries. During the prediction phase, a binary classifier assigns a score to each $(NE,CC)$ pair and the concept with the highest score is selected. The score is the probability of the classification label computed by the $softmax$ function.
\subsection*{Methodology}First the corpora was downloaded with manually tagged and normalized species. Then the named entities were extracted from annotation files of all the articles/abstracts and duplicate entities were discarded. For example, if an entity ``mouse" appears multiple times in several different articles, we consider it only once for training and testing. After deduplication, we resolve acronyms manually by searching the given identifier in the NCBI taxonomy. 
\begin{center}
	\begin{figure}[h!] 
		\centering
		\includegraphics[width=12cm, height=5cm]{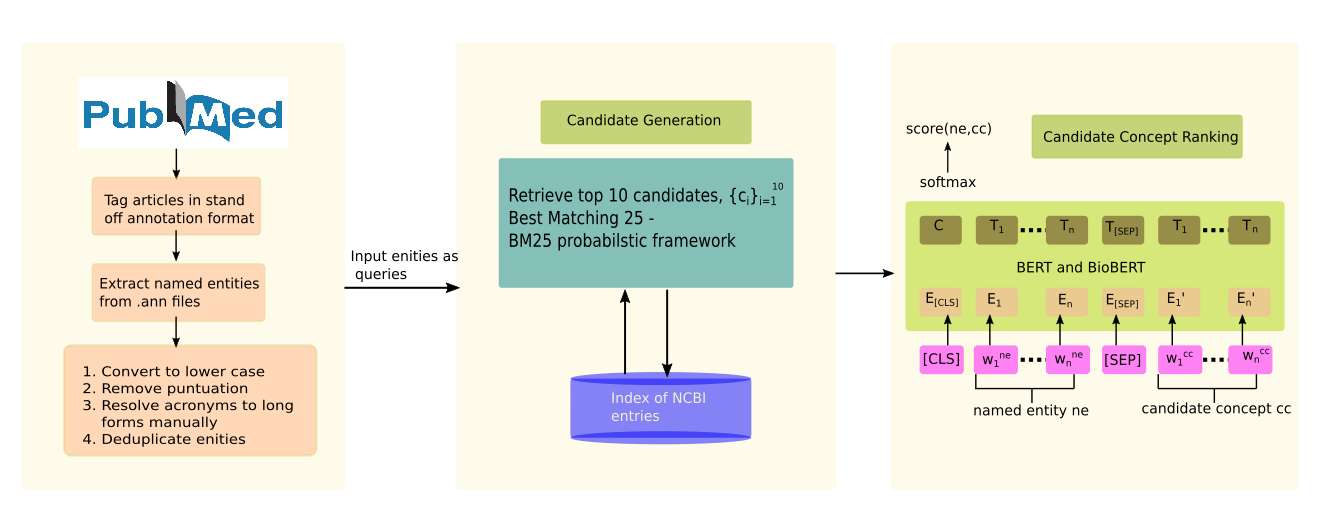}
		\caption{Normalization framework for  linking species with NCBI taxonomy identifiers}
		\label{bm25-bert}
	\end{figure}
\end{center}

Figure \ref{bm25-bert} shows the architecture used for species named entity normalization. Our method has three main steps; pre-processing, candidate generation and candidate ranking. 

\begin{center}
	\begin{figure}[h!] 
		\centering
		\includegraphics[width=10cm, height=6cm]{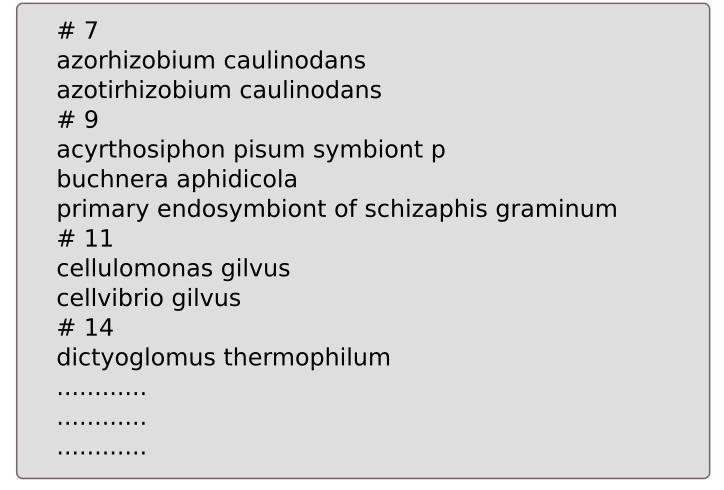}
		\caption{A snippet of NCBI taxonomy transformed to a corpus for BM25}\label{ncbidict}
	\end{figure}
\end{center}

\paragraph{\textbf{Pre-Processing}}
In the first step, we extract the named entities from .ann files and apply the following pre-processing steps. We convert them to lower case ASCII, remove punctuation marks and resolve acronyms to long-form manually. These named entities (species) will serve as queries to the BM25 algorithm \cite{loper2002nltk}. We use the spaCy toolkit\footnote{https://spacy.io/} to pre-process documents.
\paragraph{\textbf{NCBI Taxonomy Dictionary}}
We downloaded names and node files from the NCBI Taxonomy website\footnote{ftp://ftp.ncbi.nlm.nih.gov/pub/taxonomy/} on 11 March, 2020. We pre-process the names file based on the information from the nodes file. We created separate dictionaries for strains, species, phylum, order, family and genus. The identifiers of strains, species, family, genus and phylum serve as an index in the dictionary, and the actual name/phrase serves as the content of that particular index. For instance, species dictionary index $\# 7$ contains \texttt{azorhizobium caulinodans, azotirhizobium caulinodans} as its contents as shown in Figure \ref{ncbidict}.

\paragraph{\textbf{Candidate Generation}}
In this step, we give queries (species) as input to the BM25 algorithm and retrieve top ${k=10,3}$ candidates for each query. Okapi BM25, a commonly used information retrieval algorithm by search engines \cite{taylor2006optimisation,liu2011learning,robertson1995okapi}, is a probabilistic framework which retrieves a ranked list of documents for a given query. We choose the default values of parameters as $k1 = 1.2$, $k2 = 100$ and $b = 0.75$ \cite{robertson2009probabilistic}.  
\paragraph{\textbf{Candidate Ranking}}
Reranking of the retrieved list is treated as a sentence-pair classification task. A query and its candidates make sentence pairs. Each query has at most ten query-candidate pairs. An example of such a pair is shown in Table \ref{neccpair}, where \texttt{aspergillus nidulans} is the named entity that serves as the query to the BM25 algorithm which in turn retrieves ${k=10}$ candidate concepts from the NCBI Taxonomy. The objective is to rerank the list and bring the correct candidate to the top of the list. The \texttt{Label} column is $1$ if the retrieved candidate is correct and $0$ if the retrieved candidate is incorrect. The sentence classification task will maximize semantic equivalence between query and candidate concept. We use pre-trained BioBERT \cite{lee2020biobert} and Bert-base-uncased \footnote{https://huggingface.co/bert-base-uncased} for fine-tuning on the LINNAEUS and S800 corpora. The probability of $label=1$ is used as the score for reranking the list. The pair with the highest probability(score) is assigned to the input entity, where identifier = arg max $P(label=1|NE,CC)$.

\begin{table}[h!]
	\caption{Named entity (Query)- Candidate Concepts pairs example}
    \label{neccpair}
	\centering
	\renewcommand{\arraystretch}{1.5}
	\begin{tabular}{p{6cm}|p{4cm}|l}
		\hline\textbf{Candidate Concept} &  \textbf{Query Identifier} &\textbf{Label}\\ \hline
		
		aspergillus nidulans  aspergillus nidulellus  emericella nidulans (162425)	&aspergillus nidulans (162425)&	1\\ 
		aspergillus latus  aspergillus nidulans var  latus  aspergillus sp  ajc 2016b  emericella nidulans var  lata (41734)	&aspergillus nidulans (162425)	&0\\ 
		aspergillus delacroixii  aspergillus delacroxii  aspergillus nidulans var  echinulatus  aspergillus spinulosporus  emericella echinulata  emericella nidulans var  echinulata (1810908)	&aspergillus nidulans (162425)&	0 \\ 
		synechococcus nidulans(463277)	&aspergillus nidulans (162425)	&0\\ 
		mecopus nidulans (1898863)	&aspergillus nidulans (162425) &	0\\ 
		phyllotopsis nidulans(38812)	&aspergillus nidulans (162425)	&0\\ 
		nassella nidulans (523898)	&aspergillus nidulans (162425)&	0\\
		aphanothece nidulans (202207)	&aspergillus nidulans (162425)&	0 \\ 
		olgaea nidulans	(591996)&aspergillus nidulans	(162425)&0\\
		oxalis nidulans (245251)	&aspergillus nidulans (162425)&	0\\ \hline

	\end{tabular}
	
\end{table}

\subsection*{Corpora} In this section, we explain the two corpora used for evaluation purposes. 
\paragraph{\textbf{LINNAEUS}} The Linnaeus corpus \cite{gerner2010linnaeus} has $100$ full text articles randomly chosen from Pubmed Central Open Access (PMC-OA) subset. The corpus has been annotated manually for species and has been mapped to NCBI taxonomy identifiers.  
\paragraph{\textbf{S800}} The S800 corpus \cite{pafilis2013species} has 800 abstracts from PubMed from diverse taxonomic classes. It has eight taxonomic units from bacteriology, botany, entomology, medicine, mycology, protistology, virology and zoology. The corpus is manually annotated for species and normalized to NCBI taxonomy identifiers.
\section*{Experimental Setup}\label{experimentalsetup}

In this section, we discuss how the data has been split for training and evaluation and the configurations of the BERT network used for pair-wise learning to rank.

Table \ref{corporastatistics} shows that we have used LINNAEUS and  S800 corpora for training and evaluation where $80\% -10\%-10\%$ of the documents go into training, development and test subsets, respectively. Each set had duplicates of named entities, as a named entity may appear more than once in an abstract or full-text document. Hence, we applied a deduplication script to get rid of duplicate named entities. For instance, if an entity ``rat" appears multiple times in several abstracts, it is considered only once in the training subset. 


We use PyTorch's transformers library by HuggingFace\cite{wolf2019huggingface} to finetune BERT. We implement pair-wise learning to rank model by adding a linear layer with softmax activation on $[CLS]$ token. We encode the input sequences as \texttt{[CLS]named entity[SEP]candidate concept[SEP]}. We use Bert-base-uncased and BioBert \cite{lee2020biobert} pre-trained models with $10$ epochs, batch size $16$ and $3\times10^{-5}$ learning rate, and remaining hyper-parameters kept the same as in the pre-trained models. 

\begin{table}[h!]
	\caption{Corpora statistics}
	\label{corporastatistics}
	\centering
	\renewcommand{\arraystretch}{1.2}
	\begin{tabular}{p{5cm}c|c|c|c|c|cc}
		\hline
		\multirow{2}{3cm}{} & \multicolumn{3}{c|}{\textbf{LINNAEUS}} & \multicolumn{3}{c}{\textbf{S800}}\\
		\cline{2-7}
		& train & dev & test & train & dev & test\\ \hline
		
		\# of documents & 80 & 10 & 10 & 640 &80& 80 \\ \hline
		\# of unique mentions & 257 & 102 & 44 & 1016 &133&134\\ \hline
	\end{tabular}
	
\end{table}

\section*{Evaluation and Discusion}\label{resultsanddiscussions}
In this section, we evaluate our method on two corpora, LINNAEUS and S800. We report our findings and compare with OrganismTagger \cite{naderi2011organismtagger} and ORGANISMS \cite{pafilis2013species}. We use accuracy as an evaluation metric. If the correctly linked identifier is ranked first (highest score) out of $10$ candidates for a named entity, we consider it correct. We considered top-10 and top-3 candidates. When we used top-3 candidates, it resulted in a slightly lower performance. Hence, we used top-10 candidates only. Accuracy is measured as the number of correctly normalized entities divided by the total number of entities. For evaluation, we consider only those entities for which at least one correct candidate is generated in the candidate generation step. If no correct candidate is generated in the candidate-generation step, then there is no purpose of re-ranking the incorrect candidates. 

We perform five sets of experiments to evaluate our test data on OrganismTagger, ORGANISMS, BM25, BM25+Bert-base-uncased and BM25+BioBERT. OrganismTagger and ORGANISMS are the two state-of-the-art baseline methods for species normalization to the NCBI taxonomy. BM25 algorithm is used to generate candidate identifiers for the named entities(queries) in the test set. BM25+bert-base-uncased and BM25+BioBERT are used to re-rank the candidates generated from the BM25 algorithm. We report the accuracy of BM25 alone, where we consider the top-most candidate and do not consider the rest of the candidates. We then report accuracy after re-ranking the top-10 candidates generated from the BM25 generator. 

Table \ref{evaluation} shows that the BM25 generator combined with BioBERT outperforms OrganismTagger, ORGANISMS, BM25, and BM25+BERT-base-uncased. For the  LINNAEUS corpus, when the BM25 generator was used alone, we see that only $50\%$ of the entities were normalized correctly. However, after applying BERT based re-ranker, the accuracy is improved to $80\%$ and $86.66\%$ for BM25+Bert-base-uncased BM25+BioBERT methods, respectively.

For the S800 corpus, Table \ref{evaluation} shows an accuracy of $71.76\%$ for OrganismTagger and $72.41\%$ for ORGANISMS BM25+BERT-base-uncased; the number of correctly normalized entities were the same for each of these but, the entities were different.
The BM25 generator retrieves $58.6\%$ correct identifiers for entities. Re-ranking improves the accuracy from $58.6\%$ to $79.31\%$ for the BM25+BioBert model.

\begin{table}[h!]
	\caption{Evaluation on LINNAEUS and S800 corpora for the test set}
	\label{evaluation}
	\centering
	\renewcommand{\arraystretch}{1.4}
	\begin{tabular}{p{5cm}cc}
		\hline	\hline &{\textbf{LINNAEUS}} &{\textbf{S800}}\\
		\cline{2-3}
		& Accuracy& Accuracy\\ \hline	\hline
		OrganismTagger \cite{naderi2011organismtagger}& 44.57$\pm$5.12 & 78.66$\pm$4.59
		\\
		ORGANISMS \cite{pafilis2013species} &  50.74$\pm$11.2 & 74.62$\pm$3.27\\ 
		BM25 &  59.58$\pm$13.56 & 67.81$\pm$7.72 \\ 
		BM25+bert-base-uncased & 82.46$\pm$3.67&84.16$\pm$7.51\\ 
		BM25+BioBERT &88.56$\pm$5.12&  86.86$\pm$4.96\\ \hline	\hline
	\end{tabular}
	
\end{table}
One limitation of this work is that for evaluation purpose, we consider only those entities for which BM25 generates at least one correct candidate out of the top-10 candidates. We discard those named entities for which no correct candidate was generated as reranking will not help. In this case, after generating candidates, we have $30$ entities in LINNAEUS and $87$ entities in the S800 test set.

We show interesting use cases in Table \ref{usecase}, where we observe that for named entities \texttt{child\footnote{https://organisms.jensenlab.org/Search?query=child, (accessed 6 October 2020)}, children\footnote{https://organisms.jensenlab.org/Search?query=childrena, (accessed 6 October 2020)}, fire ant\footnote{https://organisms.jensenlab.org/Search?query=fire\%20ant, (accessed 6 October 2020)}} and \texttt{Asian rice\footnote{https://organisms.jensenlab.org/Search?query=Asian\%20rice, (accessed 6 October 2020)}} ORGANISMS \cite{pafilis2013species} assign identifiers based on the lexical overlap and ignores the semantic context of entities. For all of these named entities our proposed framework was able to understand semantics and hence assigned the correct identifiers. For instance, our approach assigned $9606$ to \texttt{children} instead of assigning $525814$ which is \texttt{Childrena}, a genus of butterflies.
\begin{table}[h!]
	\caption{Examples of species and their identifiers assigned by ORGANISMS \cite{pafilis2013species}}
	\label{usecase}
	\centering
	\renewcommand{\arraystretch}{1.5}
	\begin{tabular}{p{1.5cm}|c|p{4cm}c}
		\hline\textbf{Query} & \textbf{Correct Identifier }& \textbf{ ORGANISMS \cite{pafilis2013species}}\\ \hline
		
		Children & 9606 & Childrena [525814]
		
		\\ \hline
				Child & 9606 & Childia [84070]
				
				\\ \hline
		Fire ant &13686& Fire ant associated circular virus 1 [2293280]\\ \hline
			Asian rice&4530	&Orseolia oryzae [33408]
			
			\\ \hline

	\end{tabular}
	
\end{table}
\section*{Conclusion}\label{conclusion1}

In this research paper, we propose a comprehensive methodology for normalizing species entities to their corresponding NCBI taxonomy identifiers. Our approach consists of three key steps: the creation of a dictionary or corpus from the NCBI Taxonomy, candidate generation using the probabilistic information retrieval framework BM25, and re-ranking based on the powerful bi-directional encoder representation from the encoder (BERT). We evaluate the effectiveness of our proposed pipeline on two widely used benchmark corpora, LINNAEUS and S800, specifically focusing on species entities. Our experimental results demonstrate that the BERT-based re-ranking significantly improves the accuracy of normalization by assigning higher ranks to the correct identifiers within the candidate list. However, the quality of candidates generated by the BM25 framework plays a crucial role in the success of the re-ranking process. Depending on the size and nature of the ontology to which entities need to be linked, rule-based, probabilistic, and semantic search-based candidate generators can be employed. Future research directions include exploring better candidate generation techniques to fully leverage the potential of BERT-based re-ranking. Additionally, the use of a joint model based on BERT for both species recognition and normalization could be investigated to minimize error propagation from the recognition stage to the normalization stage.


\begin{backmatter}

\section*{Competing interests}
None declared.

\section*{Author's contributions}
    Text for this section \ldots

\section*{Acknowledgements}
Text for this section \ldots

\bibliographystyle{bmc-mathphys} 
\bibliography{bmc_article}      




\section*{Figures}




\section*{Additional Files}
  \subsection*{Additional file 1 --- Sample additional file title}
    Additional file descriptions text (including details of how to
    view the file, if it is in a non-standard format or the file extension).  This might
    refer to a multi-page table or a figure.

  \subsection*{Additional file 2 --- Sample additional file title}
    Additional file descriptions text.

\end{backmatter}
\end{document}